\documentclass[conference]{IEEEtran}
\ifCLASSINFOpdf
\else
\fi
\usepackage{graphicx}
\usepackage{multirow}

\begin{document}
%
\title{AI Transparency Atlas: Framework, Scoring, and Real-Time Model Card Evaluation Pipeline}

\author{
  \IEEEauthorblockN{
    Akhmadillo Mamirov\IEEEauthorrefmark{1},
    Faiaz Azmain\IEEEauthorrefmark{1},
    Hanyu Wang\IEEEauthorrefmark{2}
  }
  \IEEEauthorblockA{
    \IEEEauthorrefmark{1}Department of Computer Science, The College of Wooster, Wooster, OH, USA \\
    Emails: amamirov26@wooster.edu, fazmain25@wooster.edu
  }
  \IEEEauthorblockA{
    \IEEEauthorrefmark{2}Robert F. Wagner Graduate School of Public Service, New York University, New York, NY, USA \\
    Email: hw3592@nyu.edu
  }
}


%


\maketitle

\begin{abstract}
AI model documentation is fragmented across platforms and inconsistent in structure, preventing policymakers, auditors, and users from reliably assessing safety claims, data provenance, and version changes. We analyzed documentation from five frontier models (Gemini 3, Grok 4.1, Llama 4, GPT-5, Claude 4.5) and 100 Hugging Face model cards, identifying 947 unique section names with extreme naming variation—usage information alone appeared under 97 different labels. Using the EU AI Act Annex IV and Stanford Transparency Index as baselines, we developed a weighted transparency framework with 8 sections and 23 subsections that prioritizes safety-critical disclosures (Safety Evaluation: 25\%, Critical Risk: 20\%) over technical specifications. We implemented an automated multi-agent pipeline that extracts documentation from public sources and scores completeness through LLM consensus. Evaluating 50 models across vision, multimodal, open-source, and closed-source systems cost less than \$3 total and revealed systematic gaps: frontier labs (xAI, Microsoft, Anthropic) achieve ~80\% compliance, while most providers fall below 60\%. Safety-critical categories show the largest deficits—deception behaviors, hallucinations, and child safety evaluations account for 148, 124, and 116 aggregate points lost respectively across all evaluated models.
\end{abstract}


%
\IEEEpeerreviewmaketitle

\section{Introduction}
AI model documentation today is fragmented across whitepapers, GitHub READMEs, Hugging Face model cards, system cards, and blog posts. This fragmentation raises a core question: what practical steps can move the ecosystem from documentation inconsistency toward something standardized enough to be useful?

Documentation gaps affect every stakeholder. Regulators cannot reliably assess governance or safety without consistent reporting. Downstream institutions such as hospitals, schools, and public agencies lack visibility into model risks, evaluation protocols, and version-level changes. Even within the same platform, model cards differ dramatically in length, scope, and granularity. Some include details about architecture, training data, and evaluation settings, while others provide only brief paragraphs. Critically, documentation often does not evolve as models evolve. Major capability updates, training adjustments, and safety interventions rarely trigger corresponding updates. Versioning is ad hoc or entirely absent, causing transparency to degrade over time.

System cards attempt to address transparency at the deployment level, but they introduce their own challenges. Many system cards are high-level but not actionable; others remain closed, incomplete, or not understandable to external auditors. When modern AI systems depend on chains of interconnected models, datasets, and processes, opacity at the system layer becomes a structural barrier to accountability. When something goes wrong, responding is difficult because relevant information is scattered across multiple documents and repositories ~\cite{winecoff2024improvinggovernanceoutcomesai}.

\subsection{Why Does This Inconsistency Persist?}
The persistence of fragmented AI documentation reflects structural challenges in the AI development ecosystem. Unlike regulated industries where documentation is mandatory and enforcement mechanisms are well established, AI transparency remains largely voluntary, resulting in misaligned incentives.

\textbf{Economic and competitive pressures.}
Comprehensive documentation is resource-intensive, requiring dedicated teams for safety evaluations, data provenance tracking, and version management. Organizations prioritizing rapid deployment often treat documentation as secondary to product development. Closed-source developers face additional tension: detailed transparency can expose competitive advantages related to training methods, data sources, or architectural choices. As a result, disclosure decisions frequently involve trade-offs between transparency commitments and intellectual property protection.

\textbf{Organizational fragmentation.}
High-quality documentation requires coordination across teams that typically operate independently. Engineers prioritize model performance, safety teams focus on risk assessment, and communications teams manage external messaging. Without integrated workflows that treat documentation as a natural byproduct of development, information remains siloed across internal wikis, isolated reports, and fragmented public communications.

\textbf{Competing standards.}
Developers encounter overlapping documentation proposals, including Model Cards, Datasheets for Datasets, System Cards, and emerging regulatory frameworks, each emphasizing different priorities and formats. In the absence of a unified standard specifying required content, level of detail, and structure, developers make inconsistent choices. Some release brief model cards, while others publish extensive technical reports. Neither approach consistently satisfies stakeholder needs because no authoritative standard exists.

\textbf{Limited accountability.}
Documentation quality improves when it can be credibly evaluated. Currently, users lack practical means to assess completeness or accuracy, regulators lack scalable audit tools, and civil society organizations can identify only the most visible gaps. Without mechanisms to systematically measure and publicly compare documentation quality, providers face limited accountability for incomplete or uneven disclosures.

\subsection{Our Approach Builds on Existing Regulatory and Academic Frameworks.} 
We address these challenges through a structured transparency framework grounded in existing regulatory and academic standards. The EU AI Act Annex IV provides concrete documentation requirements that we adopt as a regulatory baseline ~\cite{eu_ai_act_annex_iv}. We use the Stanford Transparency Index as an academic reference point for evaluating disclosure completeness across AI providers \cite{bommasani2024fmti}.

 To populate this framework at scale, we built an automated pipeline that extracts documentation from dispersed public sources, evaluates completeness using multi-agent LLM consensus, and generates transparency scores. This approach enables systematic, continuous assessment of documentation quality across hundreds of models without requiring direct developer cooperation.

\section{Background and Related Work}
The Foundation Model Transparency Index (May 2024) by Stanford University provides a comprehensive assessment of foundation model developer transparency \cite{bommasani2024fmti}. The index evaluates disclosure practices across major AI labs using data provided directly by developers. While this approach offers depth and accuracy for covered models, it relies on developer cooperation and self-reported information. Our work takes a complementary approach: we evaluate documentation completeness using only publicly available information online—the same information accessible to regulators, auditors, and downstream users in practice. This distinction is critical because real-world transparency depends on what stakeholders can actually access and verify, not just what developers are willing to share upon request.

\begin{figure}[th]
    \centering
    \includegraphics[width=\linewidth]{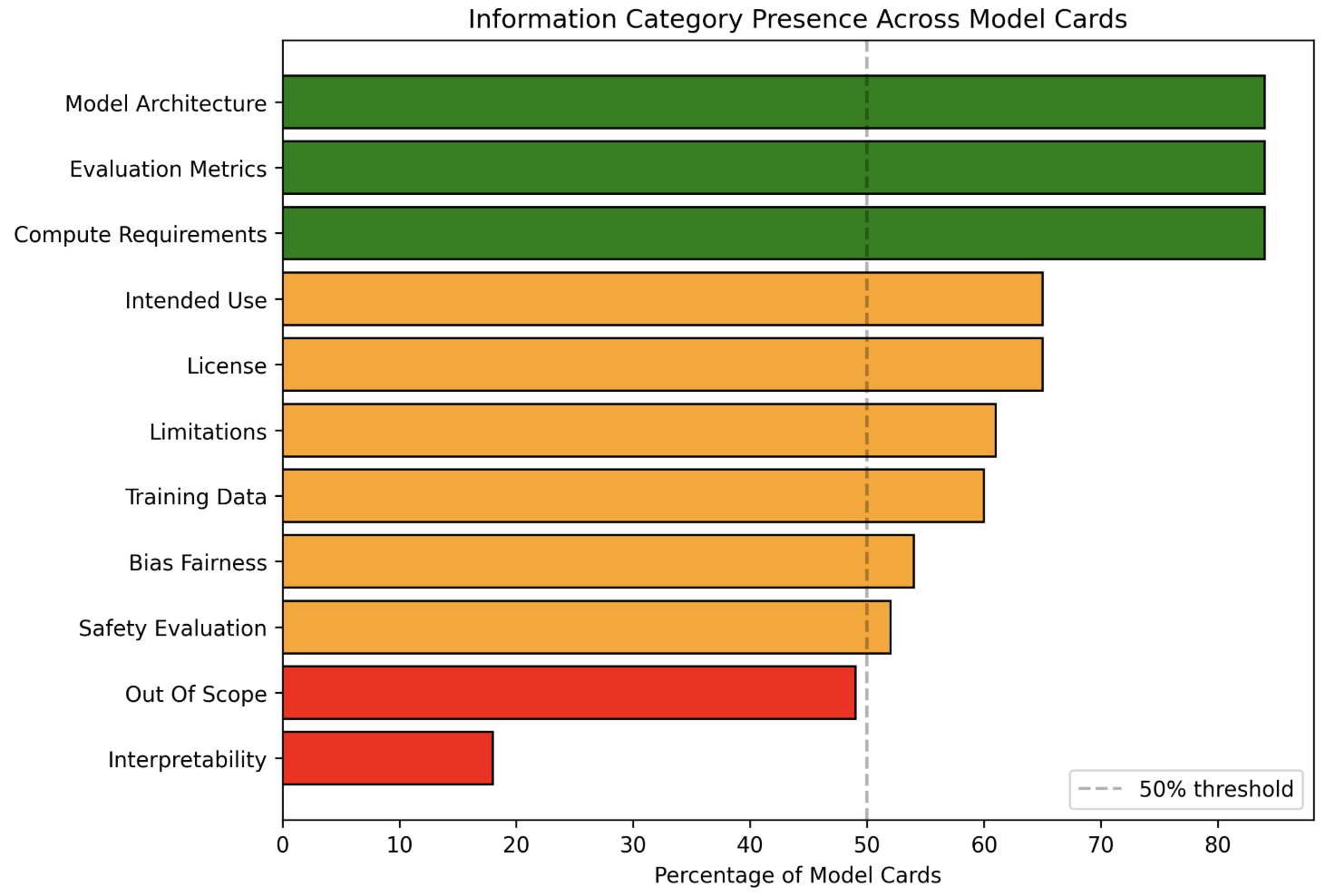}
    \caption{Category presence rates across 100 model cards. Technical categories exceed 90\% presence while safety-critical categories often fall below 50\%}
    \label{fig:category_presence}
\end{figure}

\section{CURRENT STATE OF AI MODEL DOCUMENTATION}
\subsection{Frontier Model Documentation Review}

To ground our framework in existing practice, we conducted a structured review of recent documentation artifacts from leading AI developers, including Google (Gemini~3), xAI (Grok~4.1), Meta (Llama~4), OpenAI (GPT\textendash5), and Anthropic (Claude~4.5). We also compared these against the original Model Cards proposal by Mitchell et al.\ (2019), which remains the conceptual baseline \cite{DBLP:journals/corr/abs-1810-03993}.

Across these models, we observed major differences in scope, depth, structure, and emphasis:

\begin{itemize}
    \item \textbf{Google’s Gemini 3} includes detailed sections on training data, architecture, sustainability, safety evaluations, and frontier-safety considerations \cite{google_gemini3_modelcard}.
    \item \textbf{xAI’s Grok 4.1} concentrates heavily on refusal behavior, adversarial robustness, dual-use risk, and transparency around data and training \cite{xai_grok4_1_modelcard}.
    \item \textbf{Meta’s Llama 4} emphasizes environmental footprint, quantization, safeguards, community governance, critical risks, and detailed fine-tuning behavior \cite{meta_llama4_modelcard}.
    \item \textbf{OpenAI’s GPT-5} provides the most extensive safety-evaluation sections, including red-teaming by multiple external organizations, biological risk assessments, cybersecurity stress tests, deception and sycophancy evaluations, and system-level protections \cite{openai_gpt5_system_card}.
    \item \textbf{Anthropic’s Claude 4.5} offers the most granular system-level reporting, covering agentic risks, alignment attempts, cyber ranges, interpretability studies, reward-hacking investigations, welfare assessments, and Responsibly Scaling Policy (RSP) aligned evaluations \cite{anthropic_claude_sonnet45_system_card}.
\end{itemize}

\subsection{Systematic Analysis of Hugging Face Model Cards}
To assess whether these documentation patterns extend beyond frontier models, we analyzed 100 model cards from Hugging Face, spanning diverse model types, parameter scales, and deployment contexts \cite{AI_documentation_analysis_2025}. This approach follows Liang et al. \cite{liang2024whatsdocumentedaisystematic}, who conducted a systematic analysis of 32,000 AI model cards and demonstrated that Hugging Face provides a representative sample of documentation practices across the broader AI ecosystem. From this broader sample, we identified 11 recurring documentation categories that appeared with varying frequency and depth: \emph{Model Architecture}, \emph{Compute Requirements}, \emph{Evaluation Metrics}, \emph{License}, \emph{Intended Use}, \emph{Training Data}, \emph{Limitations}, \emph{Bias and Fairness}, \emph{Safety Evaluation}, \emph{Out-of-Scope Use}, and \emph{Interpretability}.

Each model card was evaluated according to three levels of completeness:
\begin{itemize}
    \item \textbf{Detailed}: The category includes substantive information with specific details, metrics, or actionable content.
    \item \textbf{Mentioned Only}: The category is present but limited to high-level or superficial statements.
    \item \textbf{Absent}: The category is not addressed.
\end{itemize}

Figure~\ref{fig:category_presence} illustrates category presence rates across the dataset. While over 90\% of model cards included Model Architecture, Evaluation Metrics, and Compute Requirements, several critical transparency indicators fell below the 50\% threshold. Interpretability appeared in only approximately 20\% of cards, and Out-of-Scope use cases in roughly 60\%. Even when present, Safety Evaluation (approximately 65\%) and Bias and Fairness (approximately 60\%) were often addressed superficially.

\begin{figure}[th]
    \centering
    \includegraphics[width=\linewidth]{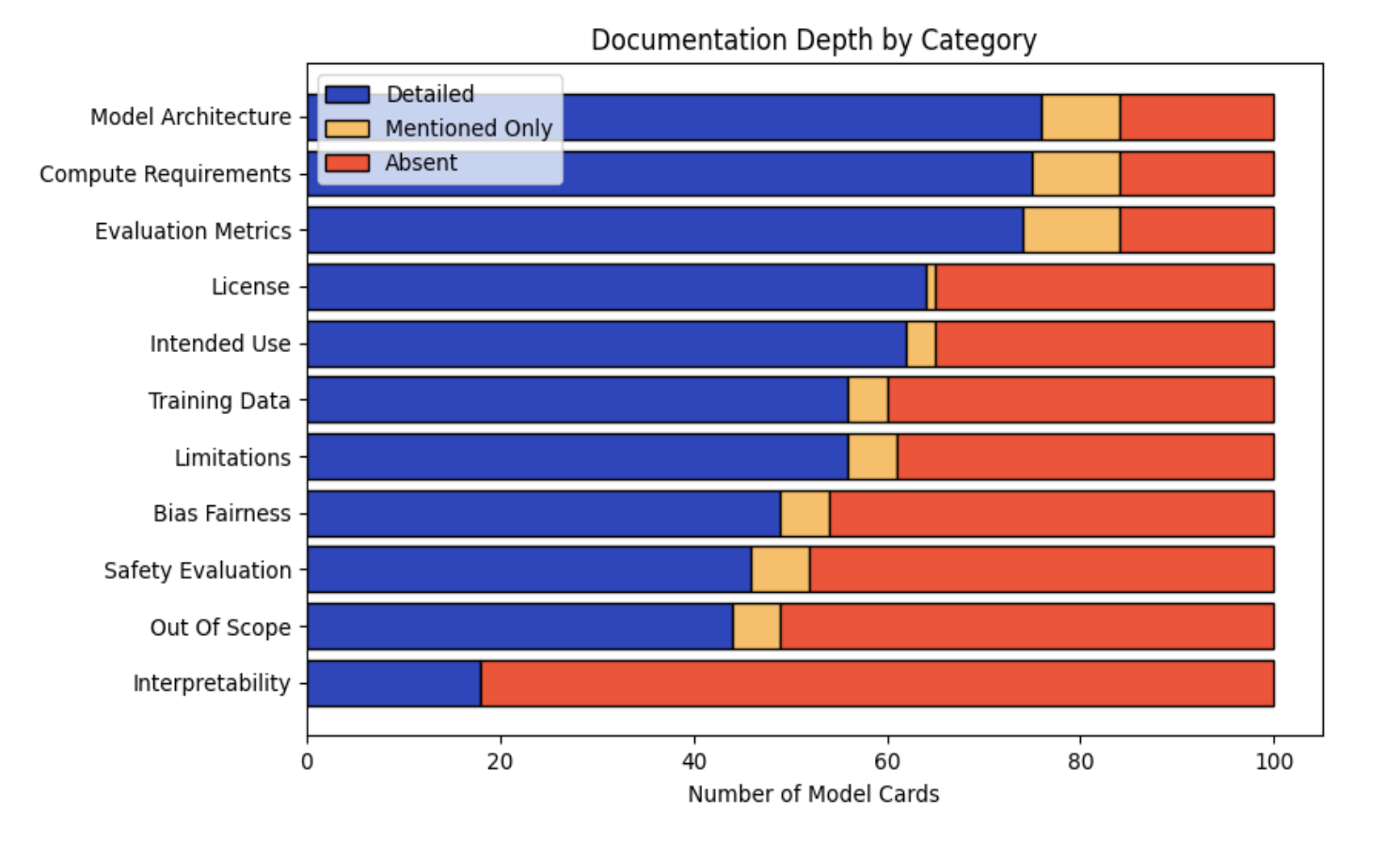}
    \caption{Documentation completeness by category across 100 Hugging Face model cards. Technical details are well-documented while safety-critical information is frequently absent.}
    \label{fig:depth_distribution}
\end{figure}

Figure~\ref{fig:depth_distribution} shows the distribution of documentation depth across all 100 model cards. Technical categories such as Model Architecture, Compute Requirements, and Evaluation Metrics were most consistently documented in detail. In contrast, safety-critical categories exhibited significant gaps: Interpretability was detailed in fewer than 20\% of model cards and absent in over 80\%. Safety Evaluation, Bias and Fairness, and Limitations were frequently mentioned but rarely described in depth.

\section{Proposed Approach}
\subsection{Defining a Minimal Core Documentation Schema}
We propose establishing a minimal consensus core schema that layers on top of existing approaches such as model cards, data sheets, and system cards. This schema would specify 20–30 essential fields that all high-impact models should report, including intended use, primary benchmarks (with exact benchmark names), high-level training data types, and known risk domains.

\begin{figure}[th]
    \centering
    \includegraphics[width=\linewidth]{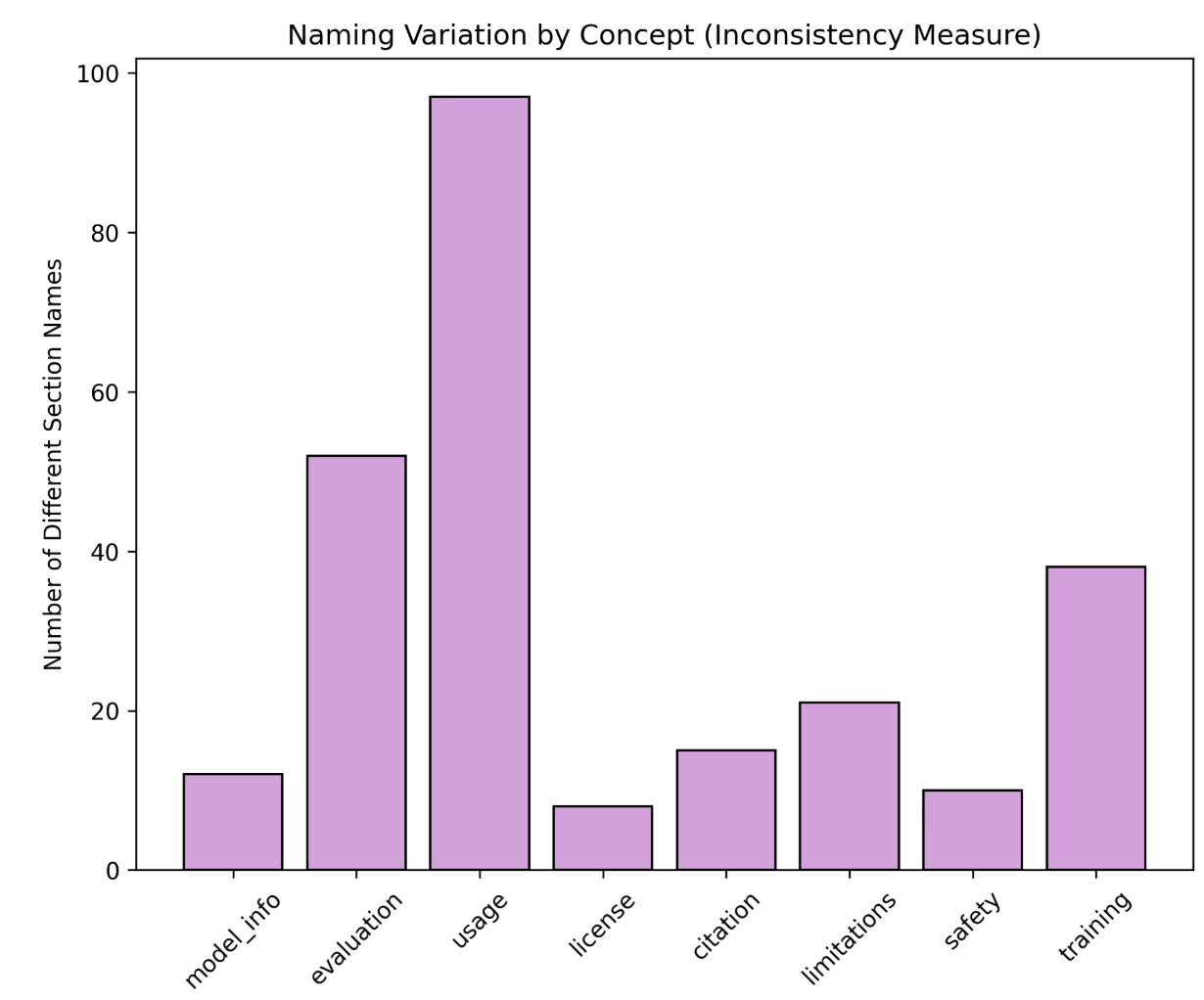}
    \caption{Naming variation across core documentation concepts. Usage information appears under 97 different section names, demonstrating severe inconsistency in model card structure}
    \label{fig:variation}
\end{figure}

Our analysis identified 947 unique section names across all cards. We used fuzzy matching to group semantically similar section names by core concept (e.g., all variations describing model usage, evaluation, or training). Figure~\ref{fig:variation}  shows the number of different section names used for each concept. Usage-related information showed the highest variation with 97 different section names, followed by evaluation (52 variations) and training (38 variations). Even fundamental concepts like license information appeared under 8 different names. This extreme variability makes it nearly impossible to systematically extract and compare information across models, reinforcing the need for a stable and shared documentation schema.

\begin{figure}[th]
    \centering
    \includegraphics[width=\linewidth]{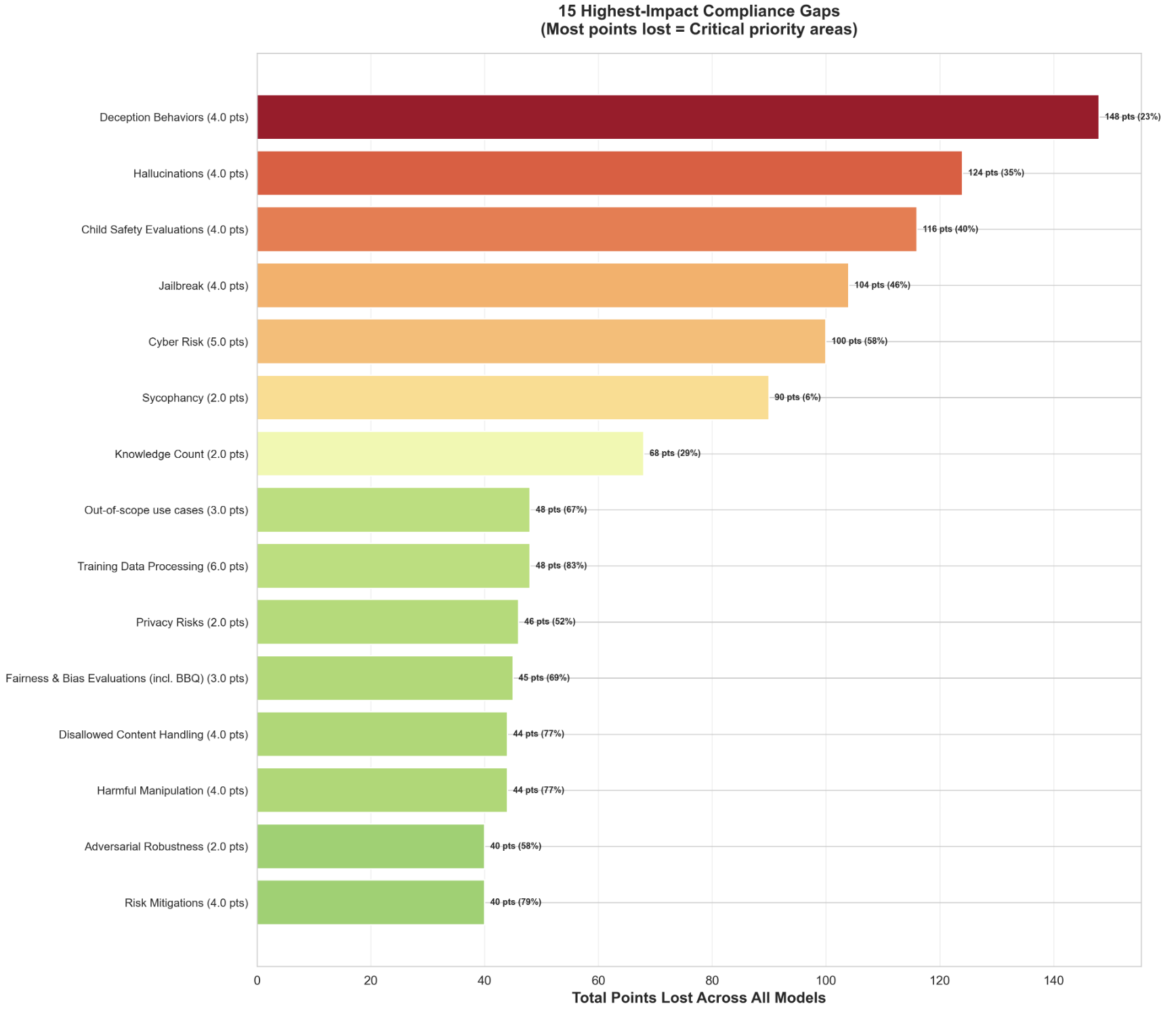}
    \caption{Aggregate point loss by subsection across all models. Higher point loss indicates categories where documentation gaps have the greatest impact on overall transparency scores due to their weight and prevalence of absence.}
    \label{fig:gaps}
\end{figure}

\begin{figure}[th]
    \centering
    \includegraphics[width=\linewidth]{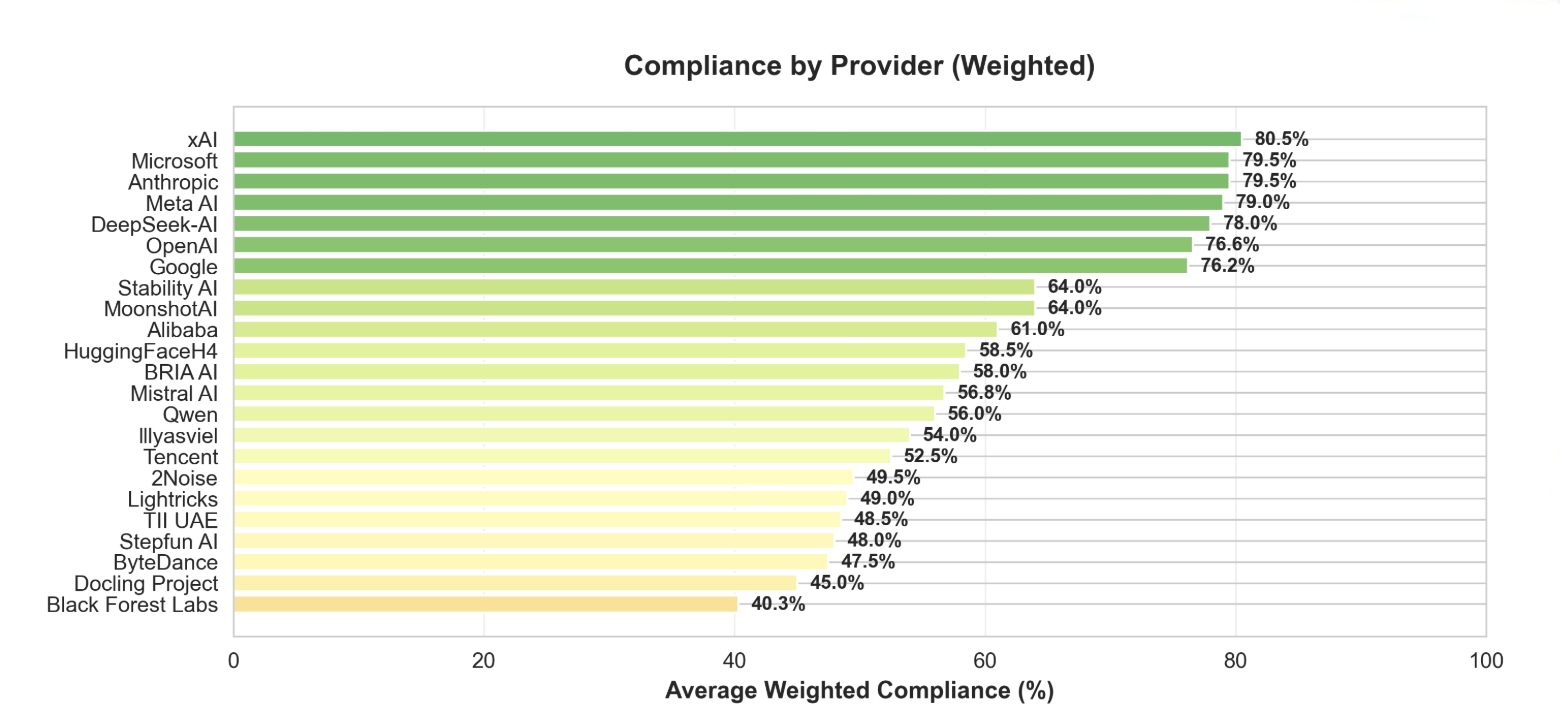}
    \caption{Weighted compliance scores by provider. Major frontier labs (xAI, Microsoft, Anthropic, Meta) achieve ~80\% compliance, while smaller providers range from 40-65\%.}
    \label{fig:compliance}
\end{figure}

We also found that:

\begin{itemize}
    \item Safety-critical fields (e.g., hallucination behavior, jailbreak resistance, cyber risk) show the largest deficits (Figure~\ref{fig:gaps}).
    \item Providers differ widely in average compliance (Figure~\ref{fig:compliance}).
    \item Weighted scoring exposes high-impact gaps that unweighted scoring fails to surface.
\end{itemize}

\subsection{Introduce a Documentation Benchmarking Scorecard}
This core schema is implemented as a framework that consists of 8 main sections and 23 subsections (see Table I). The framework synthesizes common documentation categories we identified across existing models, aligns them with EU AI Act Annex IV requirements and the Stanford Transparency Index, and prioritizes fields based on their importance for safety and governance \cite{eu_ai_act_annex_iv, bommasani2024fmti}.
Our framework assigns weighted scores to each subsection, with weights reflecting the relative importance of different disclosure types. Safety-critical information receives substantially higher weights: Safety Evaluation (25\%), Critical Risk (20\%), and Model Data (15\%) together account for 60\% of the total score, while technical specifications like Model Implementation and Sustainability (5\%) and Risk Mitigations (4\%) receive lower weights. By emphasizing safety-critical disclosures, the scorecard highlights governance-relevant gaps that unweighted scoring obscures.

\begin{table*}[t]
\centering
\begin{tabular}{l l c}
\hline
\textbf{Section} & \textbf{Subsection} & \textbf{Score (\%)} \\
\hline

\multirow{9}{*}{\textbf{Model Details (15\%)}} 
    & Model overview & 3 \\
    & Organization developing the model & 1 \\
    & Model Version & 2 \\
    & Model Release Date & 0.5 \\
    & Model Version Progression & 1 \\
    & Model Architecture & 4 \\
    & Model Dependencies & 1 \\
    & Paper and relevant links & 0.5 \\
    & Model Distribution Forms & 2 \\

\hline

\multirow{3}{*}{\textbf{Model Inputs \& Outputs (6\%)}} 
    & Inputs & 2 \\
    & Outputs & 2 \\
    & Token Count & 2 \\

\hline

\multirow{3}{*}{\textbf{Model Data (15\%)}} 
    & Training Dataset & 7 \\
    & Training Data Processing & 6 \\
    & Knowledge Count & 2 \\

\hline

\multirow{3}{*}{\textbf{Model Implementation and Sustainability (5\%)}} 
    & Hardware Used During Training \& Inference & 2 \\
    & Software Frameworks \& Tooling & 2 \\
    & Energy Use / Sustainability Metrics & 1 \\

\hline

\multirow{3}{*}{\textbf{Intended Use (10\%)}} 
    & Primary intended uses & 5 \\
    & Primary intended users & 2 \\
    & Out-of-scope use cases & 3 \\

\hline

\multirow{5}{*}{\textbf{Critical Risk (20\%)}} 
    & CBRN (Chemical, Biological, Radiological or Nuclear) & 5 \\
    & Cyber Risk & 5 \\
    & Harmful Manipulation & 4 \\
    & Child Safety Evaluations & 4 \\
    & Privacy Risks & 2 \\

\hline

\multirow{9}{*}{\textbf{Safety Evaluation (25\%)}} 
    & Refusals & 1 \\
    & Disallowed Content Handling & 4 \\
    & Sycophancy & 2 \\
    & Jailbreak & 4 \\
    & Hallucinations & 4 \\
    & Deception Behaviors & 4 \\
    & Fairness \& Bias Evaluations (incl.\ BBQ) & 3 \\
    & Adversarial Robustness & 2 \\
    & Red Teaming Results & 1 \\

\hline

\multirow{1}{*}{\textbf{Risk Mitigations (4\%)}} 
    & Risk Mitigation & 4 \\

\hline
\end{tabular}
\vspace{0.5em}
\caption{Proposed Documentation Transparency Framework: Sections, Subsections, and Scores}
\end{table*}

\subsection{Framework Objectives and Weighting Approach}
Our framework targets the two largest deficiencies in current AI documentation ecosystems: consistency and evaluability. Rather than adding yet another lengthy template, it distills the essential fields that matter for oversight and aligns them with existing regulatory expectations, making adoption practical for developers and meaningful for policymakers.

Our weighted scoring method also addresses the problem of treating all documentation fields as equally important. By weighting safety-critical disclosures more heavily, the scorecard highlights the gaps that matter most for governance. In our evaluation, this method surfaced high-risk blind spots that were invisible under unweighted scoring, producing a clearer and more policy-relevant picture of model transparency.

We recommend requiring AI developers to publish model and system documentation not only as PDFs or web pages but also as machine-readable JSON following an open, extensible schema. This schema would encode core fields such as model name and version, training data summary, evaluation benchmarks, known limitations, and implemented safety mitigations. Machine-readable documentation enables researchers, regulators, and civil-society organizations to programmatically ingest, compare, and audit models at scale.

In developing our methodology, we also examined several plausible approaches for improving AI documentation consistency and transparency. Each alternative offers conceptual advantages but also reveals structural limitations that constrain its practical effectiveness.

\begin{itemize}
    \item \textbf{Developing a comprehensive new documentation framework.}\\
    One approach is to design a fully new documentation standard that supersedes existing model cards, data sheets, and system cards. While appealing in terms of conceptual clarity, this strategy adds another framework to an already crowded landscape. Creating a standalone standard risks poor adoption, limited interoperability with existing tools, and slow industry uptake. It also overlooks areas where partial convergence already exists, such as shared fields across model cards and system cards.

    \item \textbf{Strengthening voluntary narrative guidelines.}\\
    Another approach is to rely on high-level best-practice recommendations encouraging developers to ``write better model cards.'' Although lightweight and easy to disseminate, purely narrative guidance does not correct the structural issues observed in our empirical analysis: inconsistent field definitions, lack of machine-readable formats, and wide variability in documentation depth. Moreover, voluntary guidelines provide weak incentives and no mechanism for automated evaluation, limiting their ability to drive meaningful improvement.

    \item \textbf{Establishing an immediate, globally harmonized regulatory standard.}\\
    A third approach is to push for a fully harmonized international standard for AI documentation. While this could theoretically provide a unified foundation, it is politically and technically difficult to implement in the near term. Regulatory ecosystems differ across jurisdictions, and AI systems evolve faster than global regulatory consensus can form. Such an approach risks regulatory stagnation or adoption of a lowest-common-denominator standard. As a result, it provides limited near-term utility for practitioners or policymakers.
\end{itemize}

\begin{figure*}[t]
    \centering
    \includegraphics[width=\textwidth]{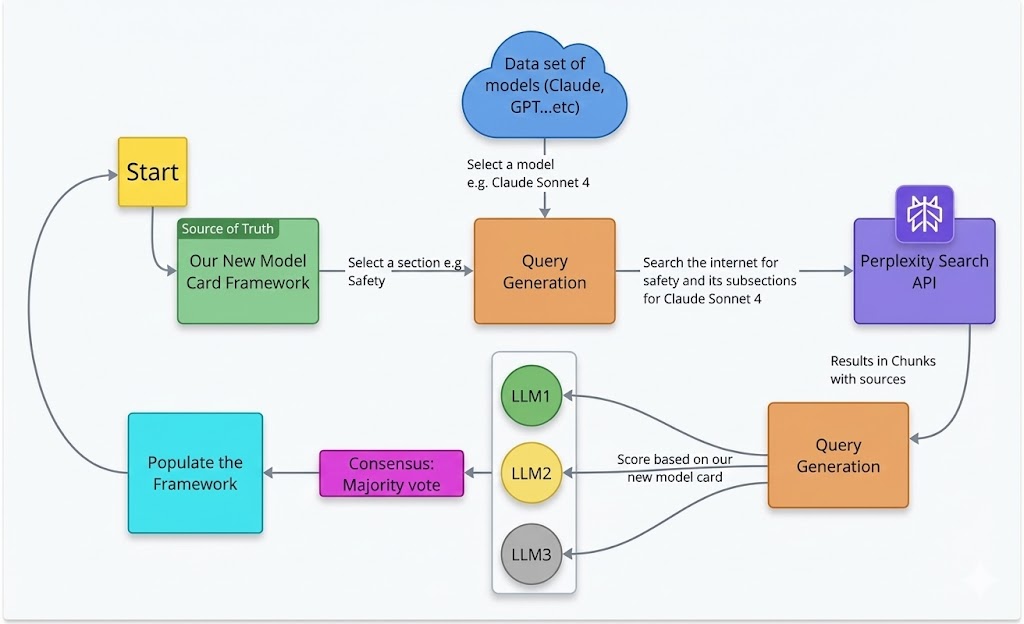}
    \caption{Overview of our automated data extraction and scoring pipeline.
    Starting from our model card framework (source of truth), the system selects a model,
    generates targeted queries for each section, retrieves evidence from the Perplexity Search API,
    evaluates the results using multiple LLMs through majority-vote consensus,
    and populates the standardized model card schema.}
    \label{fig:pipeline}
\end{figure*}

\section{Automated Documentation Extraction Pipeline}

Manual evaluation of model documentation against our framework would be time-consuming and not scalable across hundreds of models. To address this limitation, we developed an automated agentic pipeline that extracts, evaluates, and scores documentation using a multi-agent LLM system with web search capabilities \cite{AI_transparency_atlas_2025}. This approach builds on recent work by Liu et al. \cite{liu2024automaticgenerationmodeldata}, who demonstrated the feasibility of using LLMs to automatically generate model and data cards. While their work focuses on creating documentation artifacts, our pipeline evaluates and scores existing documentation by extracting information from dispersed public sources.

\subsection{Pipeline Architecture}

Figure~\ref{fig:pipeline} illustrates the overall pipeline architecture. The process begins by selecting a model from our dataset (e.g., Claude Sonnet~4, GPT-5, Llama~4). For each selected model, the pipeline iterates through the eight main sections and 23 subsections defined in our framework (see Table~I).

\subsection{Query Generation and Information Retrieval}

For each subsection (e.g., \emph{Safety Evaluation} $\rightarrow$ \emph{Jailbreak}), the pipeline generates targeted search queries designed to retrieve relevant documentation from publicly available sources. We use the Perplexity Search API to collect evidence from model cards, system cards, technical reports, blog posts, and GitHub repositories.

This design reflects real-world auditing conditions, as it evaluates only information accessible to external auditors, policymakers, and users without direct cooperation from model developers. Search results are returned in structured chunks with source citations, preserving traceability and transparency.

\subsection{Multi-Agent Consensus Scoring}

Retrieved documentation is independently evaluated by three LLM agents (LLM$_1$, LLM$_2$, LLM$_3$). Each agent assesses documentation completeness for a given subsection according to our framework criteria and assigns one of the following labels:
\begin{itemize}
    \item \textbf{Detailed}: Substantive, specific, and actionable information
    \item \textbf{Mentioned}: Present but superficial or vague information
    \item \textbf{Absent}: No relevant information found
\end{itemize}

The final score for each subsection is determined using majority-vote consensus across the three agents.

\subsection{Framework Population}

After all subsections are evaluated, the pipeline populates the framework with the extracted evidence and consensus scores. Subsection scores are then aggregated using the weighting scheme defined in Table~I to produce an overall transparency rating for each model.

\subsection{Scalability and Cost}

The pipeline is fully automated and cost-efficient. Evaluating all 50 models across 23 subsections cost less than \$3 total (under \$0.06 per model), enabling large-scale analysis across hundreds of models. We validated the pipeline on models spanning vision, multimodal, open-source, and closed-source systems, demonstrating robustness across diverse documentation formats and styles.

\section{Discussion, Limitations, and Future Work}

Our framework enables model providers to assess their documentation transparency in real-time. Providers can access a live dashboard showing their overall transparency score and subsection-level breakdowns, immediately identifying gaps in their documentation. For example, a provider scoring high on technical specifications (Model Architecture, Compute Requirements) but low on safety-critical categories (Jailbreak Evaluations, Child Safety) can prioritize improvements in those areas.

Our pipeline offers a pragmatic middle ground: instead of requiring providers to restructure their documentation, we automatically aggregate information from dispersed sources and present it through a standardized evaluation framework. This enables consistent cross-model comparison even when underlying documentation practices vary, lowering adoption barriers while improving transparency assessment for regulators, researchers, and users.

A potential limitation of any weighted scoring system is that providers may optimize for score maximization rather than genuine transparency. Since our framework assigns higher weights to safety-critical categories (Safety Evaluation: 25\%, Critical Risk: 20\%), providers could strategically prioritize these sections to boost scores while neglecting lower-weighted but still important disclosures.

However, several design features mitigate this risk. First, our framework requires substantive, detailed information—not just the presence of a section. Surface-level statements that mention safety evaluations without specific methodologies, results, or limitations receive lower scores than comprehensive disclosures. Second, the multi-agent consensus mechanism evaluates documentation quality, not just existence, making it harder to game through minimal compliance. Third, transparency scores are most valuable when used comparatively and longitudinally; providers who inflate scores through selective disclosure will be evident when their documentation is examined alongside peers or tracked over time.

Beyond point-in-time evaluation, our structured representation of publicly available documentation enables systematic tracking of documentation changes across model versions. By normalizing dispersed online disclosures into a consistent schema, the framework creates a stable reference that can be compared longitudinally as models evolve. While the current study focuses on static snapshots, we plan to extend this pipeline to support version-aware analysis, allowing transparency scores and subsection-level disclosures to be tracked across releases. This would enable stakeholders to monitor how safety claims, evaluation practices, and risk disclosures change over time, and to identify regressions or improvements in transparency as models are updated.

\section{Conclusion}

AI model documentation today is fragmented, inconsistent, and insufficient for meaningful oversight. Information is scattered across platforms, section naming varies wildly across providers, and safety-critical disclosures are frequently absent or superficial. These gaps undermine the ability of regulators, researchers, and downstream users to assess model risks, compare alternatives, and make informed decisions. This paper presented a structured response to these challenges. Through empirical analysis of frontier model documentation and 100 Hugging Face model cards, we identified systematic inconsistencies in how models are documented. We developed a transparency framework with 8 sections and 23 subsections, grounded in the EU AI Act Annex IV and the Stanford Transparency Index, with weighted scoring that prioritizes safety-critical information over technical specifications.To operationalize this framework at scale, we built an automated agentic pipeline that extracts documentation from public sources, evaluates completeness using multi-agent consensus, and generates transparency scores. The pipeline cost less than \$3 to evaluate 50 diverse models, demonstrating economic feasibility for continuous, large-scale monitoring. Our evaluation reveals significant transparency gaps across the ecosystem. While frontier labs like xAI, Microsoft, and Anthropic achieve ~80\% compliance, many smaller providers fall below 50\%. Categories like Interpretability and Safety Evaluation—critical for governance—remain poorly documented across most models. The framework offers practical value for multiple stakeholders. Providers can use live dashboards to identify documentation gaps and track improvements over time. Regulators gain evidence-based tools for compliance assessment without manual audits. Researchers obtain standardized metrics for cross-model comparison. Users gain visibility into which models have comprehensive safety documentation. Moving forward, transparency in AI cannot rely solely on voluntary adoption of documentation standards. Our approach demonstrates that automated extraction and standardized evaluation can bridge the gap between current fragmented practices and the structured information needed for accountability. As AI systems become more capable and widely deployed, robust documentation transparency is not optional—it is foundational to responsible governance.



%
\nocite{*}
\bibliography{main}
\bibliographystyle{IEEEtran}

\end{document}